\documentclass{article} 
\usepackage{tccml_iclr2025_conference,times}

\usepackage{amsmath,amsfonts,bm}

\def\eqref#1{equation~\ref{#1}}

\def\1{\bm{1}}

\DeclareMathAlphabet{\mathsfit}{\encodingdefault}{\sfdefault}{m}{sl}
\SetMathAlphabet{\mathsfit}{bold}{\encodingdefault}{\sfdefault}{bx}{n}

\usepackage[utf8]{inputenc} 
\usepackage[T1]{fontenc}    
\usepackage{hyperref}       
\usepackage{url}            
\usepackage{booktabs}       
\usepackage{amsfonts}       
\usepackage{nicefrac}       
\usepackage{microtype}      
\usepackage{xcolor}         
\usepackage{blindtext}
\usepackage{ circuitikz }
\usepackage{tikz}
\usepackage{graphicx}
\usepackage{tabu}
\usepackage[]{caption}

\title{Using Reinforcement Learning to Integrate Subjective Wellbeing into Climate Adaptation \\Decision Making}

\author{%
    Arthur Vandervoort$^{1,}$\thanks{Equal contribution.},
    Miguel Costa$^{1,*}$,
    Morten W. Petersen$^{1}$,
    Martin Drews$^{1}$,\\
    \textbf{Sonja Haustein$^{1}$},
    \textbf{Karyn Morrissey$^{2}$},
    \textbf{Francisco C. Pereira$^{1}$} \\
    $^{1}$Department of Technology, Management and Economics, 
    Technical University of Denmark\\
    \hspace{.55em}2800, Kgs. Lyngby, Denmark\\ 
    \hspace{.55em}\texttt{\{apiva, migcos, mwipe, mard, sonh, camara\}@dtu.dk} \\
    $^{2}$J.E. Cairnes School of Business and Economics,
    University of Galway\\
    \hspace{.55em}H91 TK33, Galway, Ireland\\
    \hspace{.55em}\texttt{karyn.morrissey@universityofgalway.ie} \\
}

\iclrfinalcopy 
\begin{document}

\maketitle

\begin{abstract}
Subjective wellbeing is a fundamental aspect of human life, influencing life expectancy and economic productivity, among others. Mobility plays a critical role in maintaining wellbeing, yet the increasing frequency and intensity of both nuisance and high-impact floods due to climate change are expected to significantly disrupt access to activities and destinations, thereby affecting overall wellbeing. Addressing climate adaptation presents a complex challenge for policymakers, who must select and implement policies from a broad set of options with varying effects while managing resource constraints and uncertain climate projections. In this work, we propose a multi-modular framework that uses reinforcement learning as a decision-support tool for climate adaptation in Copenhagen, Denmark. Our framework integrates four interconnected components: long-term rainfall projections, flood modeling, transport accessibility, and wellbeing modeling. This approach enables decision-makers to identify spatial and temporal policy interventions that help sustain or enhance subjective wellbeing over time. By modeling climate adaptation as an open-ended system, our framework provides a structured framework for exploring and evaluating adaptation policy pathways. In doing so, it supports policymakers to make informed decisions that maximize wellbeing in the long run.

\end{abstract}

\section{Introduction}
\label{sec:introduction}

Subjective wellbeing can be described as how individuals self-assess their own wellbeing and how they experience positive or negative aspects in their lives, including both reflective cognitive judgements (e.g., satisfaction with life) and emotional responses \citep{diener2018advances}. Subjective wellbeing has been associated with the maintenance of physical health and living a longer life \citep{steptoeSubjectiveWellbeingHealth2015}, greater economic productivity \citep{dimariaHappinessMattersProductivity2020}, and social cohesion \citep{delheyHappierTogetherSocial2016}. Research often divides subjective wellbeing into \textit{hedonic} wellbeing, i.e., the presence or absence of pleasures and pains, and \textit{eudaimonic} wellbeing, which relates to a more abstract notion of living a meaningful, actualised life in accordance with one's 'true' self \citep{ryanHappinessHumanPotentials2001}.


The ability to move through space underpins much of one's wellbeing. The capacity to be mobile, also called a person's \textit{motility} \citep{kaufmannMeasuringTypifyingMobility2018d}, is thought to have both a direct and indirect relationship with hedonic and eudaimonic wellbeing. Directly, because motility has value 'in and of itself' as a form of freedom, autonomy, and independence, and indirectly, because a greater capacity to be mobile can facilitate participation in activities which cause immediate hedonic pleasure or support eudaimonic self-actualisation \citep{devosTravelSubjectiveWellBeing2013}.

Climate change has direct consequences on how people move through a city. In particular, due to increases in high-impact weather events \citep{ipcc2023climate}, including more extreme rainfall and rising coastal waters \citep{olesen2014fremtidige}, urban floods are expected to increase. These can significantly disrupt social and economic activities or cause substantial physical damage to property \citep{hammond2015urban}. Impacts on transportation systems can be both direct (e.g., road deterioration, washouts, or loss of vehicle control) and indirect (e.g., travel delays, increased congestion, or reduced accessibility) \citep{li2018potential, wang2020climate, lu2022overview, shahdani2022assessing}. Taken together, these impacts have knock-on effects on individuals' motility, and consequently also on their wellbeing.


Copenhagen, our case study, is one of Denmark's social flood vulnerability hotspots \citep{prallComprehensiveApproachAssessing2024}. As rising coastal waters and increased pluvial flooding look set to severely affect the lives of Copenhagen's residents \citep{olesen2014fremtidige}, policymakers are facing the urgent need to make adaptation choices to tackle climage change impacts. Yet, decision-makers need to navigate the difficult challenge of finding which policy, or sequence of policies, together with resource constraints and uncertain climate projections, maintain or even increase individuals' wellbeing. 

At the same time, there are few frameworks which comprehensively link  climate change and wellbeing \citep{intergovernmentalpanelonclimatechangeipcc2022HealthWellbeing2023}. Here, artificial intelligence tools, if used constructively, can serve as a guiding tool to use scientific evidence for policymaking \citep{tyler2023ai}. We posit that this climate change adaptation policymaking challenge can be tackled by using reinforcement learning (RL). RL can allow for policy ambitions to be made actionable under complex and dynamic domains while also inspiring active engagement \citep{gilbert2022choicesrisksrewardreports}. By constructing an integrated framework combining climate, transport, and wellbeing models, and a shift in focus towards maximising individuals' wellbeing, we can identify which policymaking pathways have the capacity to mitigate climate change impacts.

\subsection{Objectives}

Our aim is to demonstrate how wellbeing can be incorporated into policymaking decisions for climate change. We do this by integrating climate models, climate impacts, transport, and wellbeing in a RL framework. To showcase this multi-modular framework, we tackle climate-related flooding and adaptation in Copenhagen, though it can be applied to other cities and other climate hazards, too. By accounting for multiple climate scenarios, this approach allows us to identify the best set of plausible policy pathways that maximise individual wellbeing in the long run. 


\section{Proposed Methodology}
\label{sec:methodology}
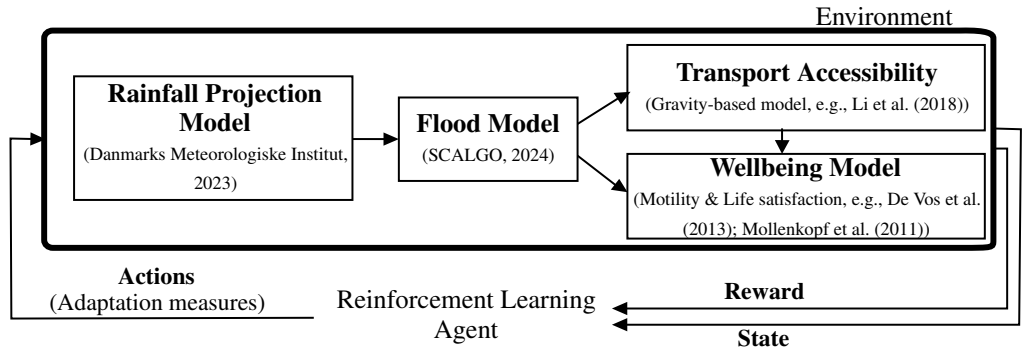
\begin{figure}[htb]
\vspace{-12pt}
\centering
    
\tikzset{every picture/.style={line width=0.75pt}} 

\begin{tikzpicture}[x=0.65pt,y=0.65pt,yscale=-.9,xscale=1]

\draw  [line width=2.25]  (70.6,25.4) .. controls (70.6,22.64) and (72.84,20.4) .. (75.6,20.4) -- (604,20.4) .. controls (620,20) and (623,22.64) .. (623,25.4) -- (623,156) .. controls (623,161) and (623,161) .. (604,161) -- (75.6,161) .. controls (72.84,161) and (70.6,158.76) .. (70.6,156) -- cycle ;
\draw    (623,96.75) -- (631,96.75) -- (631,200.25) -- (404,200.25) ;
\draw [shift={(401,200.25)}, rotate = 360] [fill={rgb, 255:red, 0; green, 0; blue, 0 }  ][line width=0.08]  [draw opacity=0] (8.93,-4.29) -- (0,0) -- (8.93,4.29) -- cycle    ;
\draw    (623,83.75) -- (639,84.25) -- (638,210.75) -- (404,210.75) ;
\draw [shift={(401,210.75)}, rotate = 360] [fill={rgb, 255:red, 0; green, 0; blue, 0 }  ][line width=0.08]  [draw opacity=0] (8.93,-4.29) -- (0,0) -- (8.93,4.29) -- cycle    ;
\draw    (227.5,206.25) -- (51.5,206.75) -- (51,90.25) -- (68.5,90.68) ;
\draw [shift={(71.5,90.75)}, rotate = 181.4] [fill={rgb, 255:red, 0; green, 0; blue, 0 }  ][line width=0.08]  [draw opacity=0] (8.93,-4.29) -- (0,0) -- (8.93,4.29) -- cycle    ;

\draw (489.25,219) node  [font=\small] [align=left] {\textbf{State}};
\draw (489.75,189) node  [font=\small] [align=left] {\textbf{Reward}};
\draw (136.25,197.5) node  [font=\small] [align=left] {\begin{minipage}[lt]{100.02pt}\setlength\topsep{0pt}
\begin{center}
{\footnotesize (Adaptation measures)}
\end{center}

\end{minipage}};
\draw (559.55,11.3) node   [align=left] {Environment};
\draw (136.25,178.9) node  [font=\small] [align=left] {\begin{minipage}[lt]{32.83pt}\setlength\topsep{0pt}
\begin{center}
\textbf{Actions}
\end{center}
\end{minipage}};

\draw    (277,63.7) -- (381,63.7) -- (381,117.7) -- (277,117.7) -- cycle  ;
\draw (329,90.7) node   [align=left] {\begin{minipage}[lt]{68pt}\setlength\topsep{0pt}
\begin{center}
\textbf{Flood Model}\\
{\scriptsize{\citep{scalgo}}}
\end{center}
\end{minipage}};

\draw    (410,29.7) -- (618,29.7) -- (618,85) -- (410,85) -- cycle  ;
\draw (514.6,58) node   [align=left] {\begin{minipage}[lt]{130pt}\setlength\topsep{0pt}
\begin{center}
\textbf{Transport Accessibility}\\
{\scriptsize{(Gravity-based model, e.g., \cite{li2018potential})}}
\end{center}
\end{minipage}};

\draw    (410,100.2) -- (618,100.2) -- (618,155) -- (410,155) -- cycle  ;
\draw (514,129) node   [align=left] {\begin{minipage}[lt]{130pt}\setlength\topsep{0pt}
\begin{center}
\textbf{Wellbeing Model}\\
{\scriptsize{(Motility \& Life satisfaction, e.g., \cite{devosTravelSubjectiveWellBeing2013, mollenkopfContinuityChangeOlder2011})}}
\end{center}
\end{minipage}};

\draw    (88,50) -- (250,50) -- (250,130) -- (88,130) -- cycle  ;
\draw (170,90) node   [align=left] {\begin{minipage}[lt]{99.08pt}\setlength\topsep{0pt}
\begin{center}
\textbf{Rainfall Projection Model}\\{\scriptsize{\citep{dmi2023klimaatlas}}}
\end{center}
\end{minipage}};

\draw  [draw opacity=0]  (225.35,178.7) -- (407.35,178.7) -- (407.35,232.7) -- (225.35,232.7) -- cycle  ;
\draw (316.35,205.7) node   [align=left] {\begin{minipage}[lt]{120.84pt}\setlength\topsep{0pt}
\begin{center}
Reinforcement Learning Agent
\end{center}

\end{minipage}};
\draw    (381,80.83) -- (405,64) ;
\draw [shift={(411.85,60)}, rotate = 150] [fill={rgb, 255:red, 0; green, 0; blue, 0 }  ][line width=0.08] [draw opacity=0] (8.93,-4.29) -- (0,0) -- (8.93,4.29) -- cycle    ;
\draw    (381,101.34) -- (405,122) ;
\draw [shift={(410.1,125)}, rotate = 215] [fill={rgb, 255:red, 0; green, 0; blue, 0 }  ][line width=0.08]  [draw opacity=0] (8.93,-4.29) -- (0,0) -- (8.93,4.29) -- cycle    ;
\draw    (250,90.59) -- (275,90.62) ;
\draw [shift={(277,90.62)}, rotate = 180.08] [fill={rgb, 255:red, 0; green, 0; blue, 0 }  ][line width=0.08]  [draw opacity=0] (8.93,-4.29) -- (0,0) -- (8.93,4.29) -- cycle    ;
\draw    (500,83.7) -- (500,97.2) ;
\draw [shift={(500,100.2)}, rotate = 270.61] [fill={rgb, 255:red, 0; green, 0; blue, 0 }  ][line width=0.08]  [draw opacity=0] (8.93,-4.29) -- (0,0) -- (8.93,4.29) -- cycle    ;

\end{tikzpicture}
    \vspace{-12pt}
    \caption{Integrated assessment model using reinforcement learning to learn what the best set of policies are that maximize wellbeing.}
    \label{fig:iam_framework}
\end{figure}

\textbf{\textsc{Modeling Framework}}\hspace{1em}
We propose to develop an integrated assessment model (IAM) to identify policy-relevant measures that maximise long-term wellbeing. IAMs provide simplified representations of complex physical and social systems and allow for assessing  climate change responses and inherent uncertainties \citep{parson1997integrated}. While traditional IAMs primarily focus on physical or economic impacts, our model emphasizes social impacts, particularly wellbeing, and uses RL to determine optimal policies under uncertain climate scenarios. As shown in Figure \ref{fig:iam_framework}, we structure our environment as four interconnected components, which we overview below. A more detailed description used in our preliminary experiments can be found in Appendix \ref{app:a}.

\textbf{\textsc{Rainfall and Flood Modeling}}\hspace{1em}
We focus on climate change's impacts on increased intensity and frequency of rainfall events. We use Klimaatlas \citep{dmi2023klimaatlas} to model rainfall. Rainfall is simulated following different projection scenarios \citep{vanvuuren2011} and range from no rainfall to intense precipitation over a short duration (e.g., cloudbursts). 

Depending on the simulated rainfall intensity at one point in time, we model its associated flood, which can range from no accumulation of water to nuisance floods, or even high-impact floods. To model how water amasses on the ground, we use SCALGO Live \citep{scalgo}. SCALGO Live is a simple interactive static event-based tool that allows for flood depths and flow direction modelling based on rainfall intensities. In essence, water accumulates over depressions and, when overflow occurs it continues downstream, allowing one to map water depth at any point over Copenhagen.

\textbf{\textsc{Transport Accessibility}}\hspace{1em}
After modeling the distribution of water in Copenhagen for a specific rainfall, we model its impacts on transport accessibility, i.e., the ease of how land-use and transport systems enable individuals to reach activities or destinations \citep{litman2009transportation}. 
Floods directly impact the speed at which you can travel, e.g., by car or foot \citep{pregnolato2017impact, finnis2008field}, and thus, they also impact accessibility. As water levels rise, travel speeds are reduced incrementally until a critical depth is reached, point at which a given segment (e.g., road) becomes untraversable. As a consequence, travel time increases because individuals travel at a reduced speed or because they need to take a different route to avoid a flooded segment. In either case, a decrease in accessibility occurs to, for example, critical infrastructure and emergency services \citep{coles2017beyond}, transport-related infrastructure \citep{li2018potential}, or everyday activities \citep{wisniewski2020vulnerability}. In our case, we model transport accessibility for different transport modes to key activities and destinations, which have been found to impact wellbeing \citep{mokhtarianSubjectiveWellbeingTravel2019a, conceicaoEffectTransportInfrastructure2023b}.

\textbf{\textsc{Wellbeing Modeling}}\hspace{1em}
Tying motility to wellbeing is a two-step process. First, as motility is an unobservable construct, we use principal component analysis to extract measures of motility based on individuals' transport access (as computed above), and survey responses to their abilities and attitudes regarding mobility \citep{kaufmannMeasuringTypifyingMobility2018d, bernierMotilityToolUncover2019}. Second, based on these components we model the relationship between motility and wellbeing. Here, we link the extracted components with a self-reported life satisfaction measure by using a cumulative link (i.e. ordinal regression) model. The outputs of this model --- and in particular a predicted life satisfaction score --- is passed to the RL's reward function.

\textbf{\textsc{Reinforcement Learning}}\hspace{1em}
Under the current climate uncertainty, we posit to learn the best set of adaptation policies that maximise wellbeing using RL. RL uses an agent-based approach to interact with the previously defined environment by taking an action and maximising a given reward function \citep{sutton2018reinforcement}. This "trial-and-error" framework can overcome traditional methods in dealing with non-linearities, learning in an interactive environment while also adhering to its constraints, and maximise a delayed reward \citep{matsuo2022deep}. At the same time, it can balance trade-offs between competing and diverse sets of actions and their rewards and allow for non-linear effects of the environment and its input uncertainties.

In this work, we propose a small set of actions that can be implemented over different areas of Copenhagen. Actions can have both short and long term effects, like increasing road drainage, permeable paving, or implementing early warning systems. Actions change the environment and directly or indirectly affect wellbeing. We define the reward function to optimize for as a metric of wellbeing, as defined above. To learn which action to perform at any time, our RL agent takes an action and collects information about the state of our digital city. Over time, the RL agent learns the best set of policies to take which maximise our long-term cumulative wellbeing. As a whole, this framework can be used as a tool to help guide policymaking for climate change adaptation.

\section{Pathway to Impact}
\label{sec:pathway_to_impact}
Our proposed approach serves as a queryable and comprehensive framework that enables policymakers to assess the long-term impacts of adaptation interventions with the goal of maximizing wellbeing. Despite the growing recognition of wellbeing as a critical factor in climate adaptation, there is a lack of integrated frameworks that explicitly incorporate wellbeing \citep{intergovernmentalpanelonclimatechangeipcc2022HealthWellbeing2023} and health co-benefits of adaptation interventions \citep{sharifiSystematicReviewHealth2021}. Due to its dynamic and open-ended nature, it is challenging to study climate adaptation policy through traditional optimization-based approaches \citep{workmanClimatePolicyDecision2021}. This challenge is particularly pressing in Copenhagen, one of Denmark’s most flood-prone areas \citep{prallComprehensiveApproachAssessing2024}, where worsening climate conditions are increasing both the severity and frequency of flood events \citep{olesen2014fremtidige}.


Our model responds to these challenges by incorporating wellbeing into climate adaptation policy simulations in a modular and scaleable way. The RL framework, meanwhile, is well-suited to tackling the kinds of open-ended systems that characterise climate adaptation policy. Finally, our approach can support policymakers to identify plausible policy pathways which maximise the wellbeing of Copenhagen residents under climate stressed and resource constrained conditions.

\bibliography{iclr2025_conference}

\begin{thebibliography}{43}
\providecommand{\natexlab}[1]{#1}
\providecommand{\url}[1]{\texttt{#1}}
\expandafter\ifx\csname urlstyle\endcsname\relax
  \providecommand{\doi}[1]{doi: #1}\else
  \providecommand{\doi}{doi: \begingroup \urlstyle{rm}\Url}\fi

\bibitem[Adhikari et~al.(2013)Adhikari, Kheir, Greve, B{\o}cher, Malone,
  Minasny, McBratney, and Greve]{adhikari2013high}
Kabindra Adhikari, Rania~Bou Kheir, Mette~B Greve, Peder~K B{\o}cher, Brendan~P
  Malone, Budiman Minasny, Alex~B McBratney, and Mogens~H Greve.
\newblock High-resolution 3-d mapping of soil texture in denmark.
\newblock \emph{Soil Science Society of America Journal}, 77\penalty0
  (3):\penalty0 860--876, 2013.

\bibitem[Bellman(1957)]{bellman1957markovian}
Richard Bellman.
\newblock A markovian decision process.
\newblock \emph{Journal of Mathematics and Mechanics}, pp.\  679--684, 1957.

\bibitem[Bernier et~al.(2019)Bernier, Gumy, Drevon, and
  Kaufmann]{bernierMotilityToolUncover2019}
{\'E}loi Bernier, Alexis Gumy, Guillaume Drevon, and Vincent Kaufmann.
\newblock Motility as a tool to uncover mobility practices.
\newblock 2019.

\bibitem[Coles et~al.(2017)Coles, Yu, Wilby, Green, and
  Herring]{coles2017beyond}
Daniel Coles, Dapeng Yu, Robert~L Wilby, Daniel Green, and Zara Herring.
\newblock Beyond ‘flood hotspots’: Modelling emergency service
  accessibility during flooding in york, uk.
\newblock \emph{Journal of hydrology}, 546:\penalty0 419--436, 2017.

\bibitem[Concei{\c c}{\~a}o et~al.(2023)Concei{\c c}{\~a}o, Monteiro, Kasraian,
  {van den Berg}, Haustein, Alves, Azevedo, and
  Miranda]{conceicaoEffectTransportInfrastructure2023b}
Marta~Aranha Concei{\c c}{\~a}o, Mayara~Moraes Monteiro, Dena Kasraian, Pauline
  {van den Berg}, Sonja Haustein, In{\^e}s Alves, Carlos~Lima Azevedo, and
  Bruno Miranda.
\newblock The effect of transport infrastructure, congestion and reliability on
  mental wellbeing: A systematic review of empirical studies.
\newblock \emph{Transport Reviews}, 43\penalty0 (2):\penalty0 264--302, March
  2023.
\newblock ISSN 0144-1647.
\newblock \doi{10.1080/01441647.2022.2100943}.

\bibitem[{Danmarks Meteorologiske Institut}(2023)]{dmi2023klimaatlas}
{Danmarks Meteorologiske Institut}.
\newblock Klimaatlas, 2023.
\newblock URL \url{https://www.dmi.dk/klima-atlas/data-i-klimaatlas}.
\newblock {Accessed: 2024-08-26}.

\bibitem[De~Vos et~al.(2013)De~Vos, Schwanen, Van~Acker, and
  Witlox]{devosTravelSubjectiveWellBeing2013}
Jonas De~Vos, Tim Schwanen, Veronique Van~Acker, and Frank Witlox.
\newblock Travel and {{Subjective Well-Being}}: {{A Focus}} on {{Findings}},
  {{Methods}} and {{Future Research Needs}}.
\newblock \emph{Transport Reviews}, 33\penalty0 (4):\penalty0 421--442, July
  2013.
\newblock ISSN 0144-1647.
\newblock \doi{10.1080/01441647.2013.815665}.

\bibitem[Delhey \& Dragolov(2016)Delhey and
  Dragolov]{delheyHappierTogetherSocial2016}
Jan Delhey and Georgi Dragolov.
\newblock Happier together. {{Social}} cohesion and subjective well-being in
  {{Europe}}.
\newblock \emph{International Journal of Psychology}, 51\penalty0 (3):\penalty0
  163--176, 2016.
\newblock ISSN 1464-066X.
\newblock \doi{10.1002/ijop.12149}.

\bibitem[Diener et~al.(2018)Diener, Oishi, and Tay]{diener2018advances}
Ed~Diener, Shigehiro Oishi, and Louis Tay.
\newblock Advances in subjective well-being research.
\newblock \emph{Nature human behaviour}, 2\penalty0 (4):\penalty0 253--260,
  2018.

\bibitem[DiMaria et~al.(2020)DiMaria, Peroni, and
  Sarracino]{dimariaHappinessMattersProductivity2020}
Charles~Henri DiMaria, Chiara Peroni, and Francesco Sarracino.
\newblock Happiness {{Matters}}: {{Productivity Gains}} from {{Subjective
  Well-Being}}.
\newblock \emph{Journal of Happiness Studies}, 21\penalty0 (1):\penalty0
  139--160, January 2020.
\newblock ISSN 1573-7780.
\newblock \doi{10.1007/s10902-019-00074-1}.

\bibitem[Espeholt et~al.(2018)Espeholt, Soyer, Munos, Simonyan, Mnih, Ward,
  Doron, Firoiu, Harley, Dunning, Legg, and Kavukcuoglu]{impala2018}
Lasse Espeholt, Hubert Soyer, Remi Munos, Karen Simonyan, Volodymir Mnih, Tom
  Ward, Yotam Doron, Vlad Firoiu, Tim Harley, Iain Dunning, Shane Legg, and
  Koray Kavukcuoglu.
\newblock Impala: Scalable distributed deep-rl with importance weighted
  actor-learner architectures.
\newblock 2018.
\newblock \doi{10.48550/arXiv.1802.01561}.

\bibitem[Finnis \& Walton(2008)Finnis and Walton]{finnis2008field}
Kirsten~K Finnis and Darren Walton.
\newblock Field observations to determine the influence of population size,
  location and individual factors on pedestrian walking speeds.
\newblock \emph{Ergonomics}, 51\penalty0 (6):\penalty0 827--842, 2008.
\newblock URL \url{https://doi.org/10.1080/00140130701812147}.

\bibitem[Gilbert et~al.(2022)Gilbert, Dean, Zick, and
  Lambert]{gilbert2022choicesrisksrewardreports}
Thomas~Krendl Gilbert, Sarah Dean, Tom Zick, and Nathan Lambert.
\newblock Choices, risks, and reward reports: Charting public policy for
  reinforcement learning systems, 2022.
\newblock URL \url{https://arxiv.org/abs/2202.05716}.

\bibitem[Hammond et~al.(2015)Hammond, Chen, Djordjevi{\'c}, Butler, and
  Mark]{hammond2015urban}
Michael~J Hammond, Albert~S Chen, Slobodan Djordjevi{\'c}, David Butler, and
  Ole Mark.
\newblock Urban flood impact assessment: A state-of-the-art review.
\newblock \emph{Urban Water Journal}, 12\penalty0 (1):\penalty0 14--29, 2015.

\bibitem[{IPCC}(2023{\natexlab{a}})]{intergovernmentalpanelonclimatechangeipcc2022HealthWellbeing2023}
{IPCC}.
\newblock 2022: {{Health}}, {{Wellbeing}}, and the {{Changing Structure}} of
  {{Communities}}.
\newblock In \emph{Climate {{Change}} 2022 -- {{Impacts}}, {{Adaptation}} and
  {{Vulnerability}}: {{Working Group II Contribution}} to the {{Sixth
  Assessment Report}} of the {{Intergovernmental Panel}} on {{Climate
  Change}}}. Cambridge University Press, 1 edition, June 2023{\natexlab{a}}.
\newblock ISBN 978-1-00-932584-4.
\newblock \doi{10.1017/9781009325844}.

\bibitem[{IPCC}(2023{\natexlab{b}})]{ipcc2023climate}
{IPCC}.
\newblock {Section 3: Long-Term Climate and Development Futures}.
\newblock In \emph{Climate Change 2023: Synthesis Report. Contribution of
  Working Groups I, II and III to the Sixth Assessment Report of the
  Intergovernmental Panel on Climate Change [Core Writing Team, H. Lee and J.
  Romero (eds.)]}, pp.\  35--115. IPCC, Geneva, Switzerland, doi:
  10.59327/IPCC/AR6-9789291691647, 2023{\natexlab{b}}.

\bibitem[Kaufmann et~al.(2018)Kaufmann, Dubois, and
  Ravalet]{kaufmannMeasuringTypifyingMobility2018d}
Vincent Kaufmann, Yann Dubois, and Emmanuel Ravalet.
\newblock Measuring and typifying mobility using motility.
\newblock \emph{Applied Mobilities}, 3\penalty0 (2):\penalty0 198--213, July
  2018.
\newblock ISSN 2380-0127.
\newblock \doi{10.1080/23800127.2017.1364540}.

\bibitem[Li et~al.(2018)Li, Kwan, Yin, Yu, and Wang]{li2018potential}
Mengya Li, Mei-Po Kwan, Jie Yin, Dapeng Yu, and Jun Wang.
\newblock The potential effect of a 100-year pluvial flood event on metro
  accessibility and ridership: A case study of central shanghai, china.
\newblock \emph{Applied Geography}, 100:\penalty0 21--29, 2018.
\newblock ISSN 0143-6228.
\newblock \doi{https://doi.org/10.1016/j.apgeog.2018.09.001}.
\newblock URL
  \url{https://www.sciencedirect.com/science/article/pii/S0143622818302716}.

\bibitem[Liang et~al.(2018)Liang, Liaw, Nishihara, Moritz, Fox, Goldberg,
  Gonzalez, Jordan, and Stoica]{rllib2018}
Eric Liang, Richard Liaw, Robert Nishihara, Philipp Moritz, Roy Fox, Ken
  Goldberg, Joseph~E. Gonzalez, Michael~I. Jordan, and Ion Stoica.
\newblock {RLlib}: Abstractions for distributed reinforcement learning.
\newblock \emph{Proceedings of the International Conference on Machine Learning
  (ICML)}, 2018.
\newblock URL \url{https://arxiv.org/pdf/1712.09381}.

\bibitem[Litman(2009)]{litman2009transportation}
Todd Litman.
\newblock Transportation cost and benefit analysis.
\newblock \emph{Victoria Transport Policy Institute}, 31:\penalty0 1--19, 2009.

\bibitem[Lu et~al.(2022)Lu, {Shun Chan}, Chen, Chan, and Gu]{lu2022overview}
Xiaohui Lu, Faith~Ka {Shun Chan}, Wei-Qiang Chen, Hing~Kai Chan, and Xinbing
  Gu.
\newblock An overview of flood-induced transport disruptions on urban streets
  and roads in chinese megacities: Lessons and future agendas.
\newblock \emph{Journal of Environmental Management}, 321:\penalty0 115991,
  2022.
\newblock ISSN 0301-4797.
\newblock \doi{https://doi.org/10.1016/j.jenvman.2022.115991}.
\newblock URL
  \url{https://www.sciencedirect.com/science/article/pii/S030147972201564X}.

\bibitem[Matsuo et~al.(2022)Matsuo, LeCun, Sahani, Precup, Silver, Sugiyama,
  Uchibe, and Morimoto]{matsuo2022deep}
Yutaka Matsuo, Yann LeCun, Maneesh Sahani, Doina Precup, David Silver, Masashi
  Sugiyama, Eiji Uchibe, and Jun Morimoto.
\newblock Deep learning, reinforcement learning, and world models.
\newblock \emph{Neural Networks}, 152:\penalty0 267--275, 2022.
\newblock ISSN 0893-6080.
\newblock \doi{https://doi.org/10.1016/j.neunet.2022.03.037}.
\newblock URL
  \url{https://www.sciencedirect.com/science/article/pii/S0893608022001150}.

\bibitem[Mokhtarian(2019)]{mokhtarianSubjectiveWellbeingTravel2019a}
Patricia~L. Mokhtarian.
\newblock Subjective well-being and travel: Retrospect and prospect.
\newblock \emph{Transportation}, 46\penalty0 (2):\penalty0 493--513, April
  2019.
\newblock ISSN 1572-9435.
\newblock \doi{10.1007/s11116-018-9935-y}.

\bibitem[Mollenkopf et~al.(2011)Mollenkopf, Hieber, and
  Wahl]{mollenkopfContinuityChangeOlder2011}
Heidrun Mollenkopf, Annette Hieber, and Hans-Werner Wahl.
\newblock Continuity and change in older adults' perceptions of out-of-home
  mobility over ten years: A qualitative--quantitative approach.
\newblock \emph{Ageing \& Society}, 31\penalty0 (5):\penalty0 782--802, July
  2011.
\newblock ISSN 1469-1779, 0144-686X.
\newblock \doi{10.1017/S0144686X10000644}.

\bibitem[Olesen et~al.(2014)Olesen, Madsen, Ludwigsen, Boberg, Christensen,
  Cappelen, Christensen, Andersen, and Christensen]{olesen2014fremtidige}
Martin Olesen, Kristine~Skovgaard Madsen, Carsten~Ankjær Ludwigsen, Fredrik
  Boberg, Tina Christensen, John Cappelen, Ole~Bøssing Christensen,
  Katrine~Krogh Andersen, and Jens~Hesselbjerg Christensen.
\newblock Fremtidige klimaforandringer i danmark (danmarks klimacenter rapport
  nr. 6 2014).
\newblock Technical report, Danmarks Meteorologiske Institut, 2014.

\bibitem[Palacios \& El-Geneidy(2022)Palacios and
  El-Geneidy]{palacios2022cumulative}
Manuel~Santana Palacios and Ahmed El-Geneidy.
\newblock Cumulative versus gravity-based accessibility measures: which one to
  use?
\newblock \emph{Findings}, 2022.

\bibitem[Parson \& Fisher-Vanden(1997)Parson and
  Fisher-Vanden]{parson1997integrated}
Edward~A Parson and Karen Fisher-Vanden.
\newblock Integrated assessment models of global climate change.
\newblock \emph{Annual Review of Energy and the Environment}, 22\penalty0
  (1):\penalty0 589--628, 1997.

\bibitem[Prall et~al.(2024)Prall, Brandt, Halvorsen, Hansen, Dahlberg, and
  Andersen]{prallComprehensiveApproachAssessing2024}
Mia~Cassidy Prall, Urs~Steiner Brandt, Nick~Schack Halvorsen, Morten~Uldal
  Hansen, Niklas Dahlberg, and Kaija~Jumppanen Andersen.
\newblock A comprehensive approach for assessing social flood vulnerability and
  social flood risk: {{The}} case of {{Denmark}}.
\newblock \emph{International Journal of Disaster Risk Reduction},
  111:\penalty0 104686, September 2024.
\newblock ISSN 2212-4209.
\newblock \doi{10.1016/j.ijdrr.2024.104686}.

\bibitem[Pregnolato et~al.(2017)Pregnolato, Ford, Wilkinson, and
  Dawson]{pregnolato2017impact}
Maria Pregnolato, Alistair Ford, Sean~M. Wilkinson, and Richard~J. Dawson.
\newblock The impact of flooding on road transport: A depth-disruption
  function.
\newblock \emph{Transportation Research Part D: Transport and Environment},
  55:\penalty0 67--81, 2017.
\newblock ISSN 1361-9209.
\newblock \doi{https://doi.org/10.1016/j.trd.2017.06.020}.
\newblock URL
  \url{https://www.sciencedirect.com/science/article/pii/S1361920916308367}.

\bibitem[Ryan \& Deci(2001)Ryan and Deci]{ryanHappinessHumanPotentials2001}
Richard~M. Ryan and Edward~L. Deci.
\newblock On {{Happiness}} and {{Human Potentials}}: {{A Review}} of
  {{Research}} on {{Hedonic}} and {{Eudaimonic Well-Being}}.
\newblock \emph{Annual Review of Psychology}, 52\penalty0 (Volume 52,
  2001):\penalty0 141--166, February 2001.
\newblock ISSN 0066-4308, 1545-2085.
\newblock \doi{10.1146/annurev.psych.52.1.141}.

\bibitem[SCALGO(2024)]{scalgo}
SCALGO.
\newblock {SCALGO Live}, 2024.
\newblock URL \url{https://scalgo.com/live/denmark}.
\newblock {Accessed: 2024-06-07}.

\bibitem[Schulman et~al.(2017)Schulman, Wolski, Dhariwal, Radford, and
  Klimov]{ppo2017}
John Schulman, Filip Wolski, Prafulla Dhariwal, Alec Radford, and Oleg Klimov.
\newblock Proximal policy optimization algorithms.
\newblock 07 2017.
\newblock \doi{10.48550/arXiv.1707.06347}.

\bibitem[Shahdani et~al.(2022)Shahdani, Santamaria-Ariza, Sousa, Coelho, and
  Matos]{shahdani2022assessing}
Fereshteh~Jafari Shahdani, Mónica Santamaria-Ariza, Hélder~S. Sousa, Mário
  Coelho, and José~C. Matos.
\newblock {Assessing Flood Indirect Impacts on Road Transport Networks Applying
  Mesoscopic Traffic Modelling: The Case Study of Santarém, Portugal}.
\newblock \emph{Applied Sciences}, 12\penalty0 (6), 2022.
\newblock ISSN 2076-3417.
\newblock \doi{10.3390/app12063076}.
\newblock URL \url{https://www.mdpi.com/2076-3417/12/6/3076}.

\bibitem[Sharifi et~al.(2021)Sharifi, Pathak, Joshi, and
  He]{sharifiSystematicReviewHealth2021}
Ayyoob Sharifi, Minal Pathak, Chaitali Joshi, and Bao-Jie He.
\newblock A systematic review of the health co-benefits of urban climate change
  adaptation.
\newblock \emph{Sustainable Cities and Society}, 74:\penalty0 103190, November
  2021.
\newblock ISSN 2210-6707.
\newblock \doi{10.1016/j.scs.2021.103190}.

\bibitem[Shliselberg et~al.(2020)Shliselberg, Givoni, and
  Kaplan]{shliselbergBehavioralFrameworkMeasuring2020}
Rebecca Shliselberg, Moshe Givoni, and Sigal Kaplan.
\newblock A behavioral framework for measuring motility: {{Linking}} past
  mobility experiences, motility and eudemonic well-being.
\newblock \emph{Transportation Research Part A: Policy and Practice},
  141:\penalty0 69--85, November 2020.
\newblock ISSN 09658564.
\newblock \doi{10.1016/j.tra.2020.09.001}.

\bibitem[Steptoe et~al.(2015)Steptoe, Deaton, and
  Stone]{steptoeSubjectiveWellbeingHealth2015}
Andrew Steptoe, Angus Deaton, and Arthur~A. Stone.
\newblock Subjective wellbeing, health, and ageing.
\newblock \emph{The Lancet}, 385\penalty0 (9968):\penalty0 640--648, February
  2015.
\newblock ISSN 0140-6736, 1474-547X.
\newblock \doi{10.1016/S0140-6736(13)61489-0}.

\bibitem[{Styrelsen for Dataforsyning og
  Infrastruktur}(2024)]{danishelevationmodel}
{Styrelsen for Dataforsyning og Infrastruktur}.
\newblock Danmarks højdemodel.
\newblock
  \url{https://sdfi.dk/data-om-danmark/vores-data/danmarks-hoejdemodel}, 2024.
\newblock Accessed: 2024-06-07.

\bibitem[Sutton \& Barto(2018)Sutton and Barto]{sutton2018reinforcement}
Richard~S Sutton and Andrew~G Barto.
\newblock \emph{Reinforcement learning: An introduction}.
\newblock MIT press, 2018.

\bibitem[Tyler et~al.(2023)Tyler, Akerlof, Allegra, Arnold, Canino, Doornenbal,
  Goldstein, Budtz~Pedersen, and Sutherland]{tyler2023ai}
Chris Tyler, KL~Akerlof, Alessandro Allegra, Zachary Arnold, Henriette Canino,
  Marius~A Doornenbal, Josh~A Goldstein, David Budtz~Pedersen, and William~J
  Sutherland.
\newblock Ai tools as science policy advisers? the potential and the pitfalls.
\newblock \emph{Nature}, 622\penalty0 (7981):\penalty0 27--30, 2023.

\bibitem[Van~Vuuren et~al.(2011)Van~Vuuren, Edmonds, Kainuma, Riahi, Thomson,
  Hibbard, Hurtt, Kram, Krey, Lamarque, et~al.]{vanvuuren2011}
Detlef~P Van~Vuuren, Jae Edmonds, Mikiko Kainuma, Keywan Riahi, Allison
  Thomson, Kathy Hibbard, George~C Hurtt, Tom Kram, Volker Krey, Jean-Francois
  Lamarque, et~al.
\newblock The representative concentration pathways: an overview.
\newblock \emph{Climatic change}, 109:\penalty0 5--31, 2011.

\bibitem[Wang et~al.(2020)Wang, Qu, Yang, Nichol, Clarke, and
  Ge]{wang2020climate}
Tianni Wang, Zhuohua Qu, Zaili Yang, Timothy Nichol, Geoff Clarke, and Ying-En
  Ge.
\newblock Climate change research on transportation systems: Climate risks,
  adaptation and planning.
\newblock \emph{Transportation Research Part D: Transport and Environment},
  88:\penalty0 102553, 2020.

\bibitem[Wi{\'s}niewski et~al.(2020)Wi{\'s}niewski, Borowska-Stefa{\'n}ska,
  Kowalski, and Sapi{\'n}ska]{wisniewski2020vulnerability}
Szymon Wi{\'s}niewski, Marta Borowska-Stefa{\'n}ska, Micha{\l} Kowalski, and
  Paulina Sapi{\'n}ska.
\newblock Vulnerability of the accessibility to grocery shopping in the event
  of flooding.
\newblock \emph{Transportation Research Part D: Transport and Environment},
  87:\penalty0 102510, 2020.

\bibitem[Workman et~al.(2021)Workman, Darch, Dooley, Lomax, Maltby, and
  Pollitt]{workmanClimatePolicyDecision2021}
Mark Workman, Geoff Darch, Kate Dooley, Guy Lomax, James Maltby, and Hector
  Pollitt.
\newblock Climate policy decision making in contexts of deep uncertainty - from
  optimisation to robustness.
\newblock \emph{Environmental Science \& Policy}, 120:\penalty0 127--137, June
  2021.
\newblock ISSN 1462-9011.
\newblock \doi{10.1016/j.envsci.2021.03.002}.

\end{thebibliography}
\bibliographystyle{iclr2025_conference}

\newpage
\appendix
\section{Comprehensive Description of the Model Components used in Preliminary Experiments}
\label{app:a}

\subsection{Rainfall Projection Model}
\begin{figure}[htb]
    \centering
    \includegraphics[trim={0cm 0cm 0cm 0cm},clip, width=.98\textwidth]{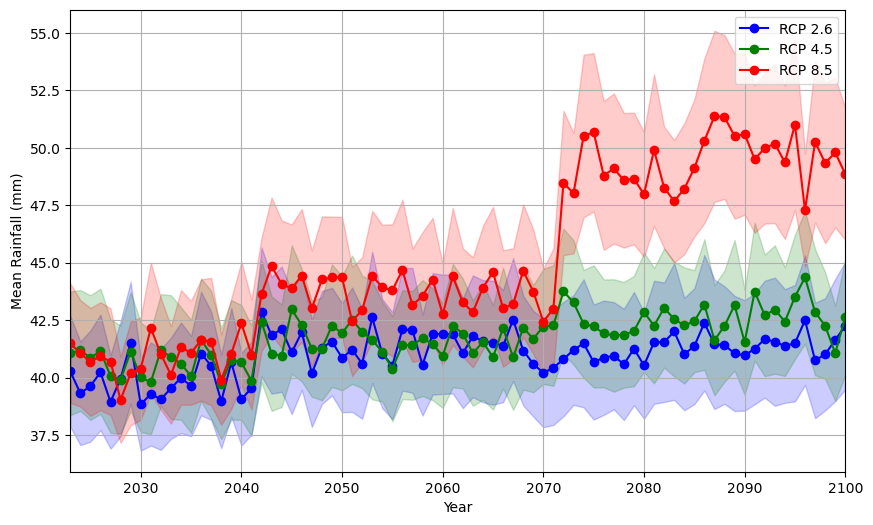}%
    \vspace{-10pt}
    \caption{Average rainfall (and 95\% confidence interval shaded) for 365 samples (equivalent to daily sampling per year) following different scenarios (RCP2.6, RCP4.5, and RCP8.5) and Climate Atlas \citep{dmi2023klimaatlas} projections for rainfall in the Copenhagen municipality between 2023 and 2100.}
    \label{fig:climate_atlas}
\end{figure}

Daily rainfall projections under different climate scenarios (e.g., RCP2.6, RCP4.5, RCP8.5) can be used to model rainfall events \citep{vanvuuren2011}. Various models exist for simulating these events, with differing levels of complexity and computational efficiency. A particularly simple and computationally efficient option is Climate Atlas, developed by the Danish Meteorological Institute, which provides stepwise projections for the periods 2011–2040, 2041–2070, and 2071–2100 \citep{dmi2023klimaatlas}. Figure~\ref{fig:climate_atlas} illustrates the average rainfall for 1,000 samples per year between 2023 and 2100, as generated from Climate Atlas.

At each time step, we simulate a daily rainfall event, assuming stationarity. We construct the cumulative density function (CDF) of Climate Atlas rainfall projections for each period and sample an event based on the selected climate scenario. The scenario is chosen at the beginning of each RL episode.

While Climate Atlas provides a straightforward, stepwise representation of rainfall increases across three periods until 2100, it has several limitations. The model assumes stationarity within each period and does not account for finer temporal variations, such as seasonal fluctuations or extreme event clustering. Additionally, its coarse spatial resolution may not capture localized rainfall dynamics, which are critical for finner urban flood modeling and adaptation planning. Future work will incorporate more advanced models that offer higher temporal and spatial granularity, dynamically represent changing climate conditions, and better integrate physical and statistical downscaling techniques. These enhancements will improve the rainfall projections, enabling more robust assessments of flood risks and adaptation strategies.

\subsection{Flood Model}

Following the simulation of a rainfall event, we model water accumulation across Copenhagen resulting from the simulated precipitation. To achieve this, we use SCALGO Live \citep{scalgo}, a computationally efficient, event-based tool for watershed delineation. SCALGO functions as a flood depth and water flow estimator, leveraging high-resolution digital terrain data to predict water accumulation and movement.

For Denmark, SCALGO integrates multiple datasets to function. It utilizes the Danish Elevation Model \citep{danishelevationmodel} for topographical information, the soil type map from the Institute of Agroecology at Aarhus University \citep{adhikari2013high} to account for soil properties such as evaporation, compaction, and vegetation cover, and sewer drainage data from the Kloakoplande (sewer plans) within the digital register for spatial planning in Denmark (\url{https://www.planinfo.dk/}) to model water runoff from artificial surfaces and surcharged drainage systems.

The model assumes a uniform rainfall distribution across all of Copenhagen with an unspecified duration, meaning rainfall intensity is consistent throughout the city, and water accumulates simultaneously across all locations. Water flow is determined by terrain properties, where runoff collects in natural depressions or low-lying areas. If the volume of accumulated water exceeds the volume capacity of a depression, it spills over and continues flowing downstream.

\begin{figure}[!ht]
    \begin{minipage}[b]{.5\textwidth}
        \centering
        \includegraphics[trim={0 2.5cm 5cm 8cm},clip, width=.99\textwidth]{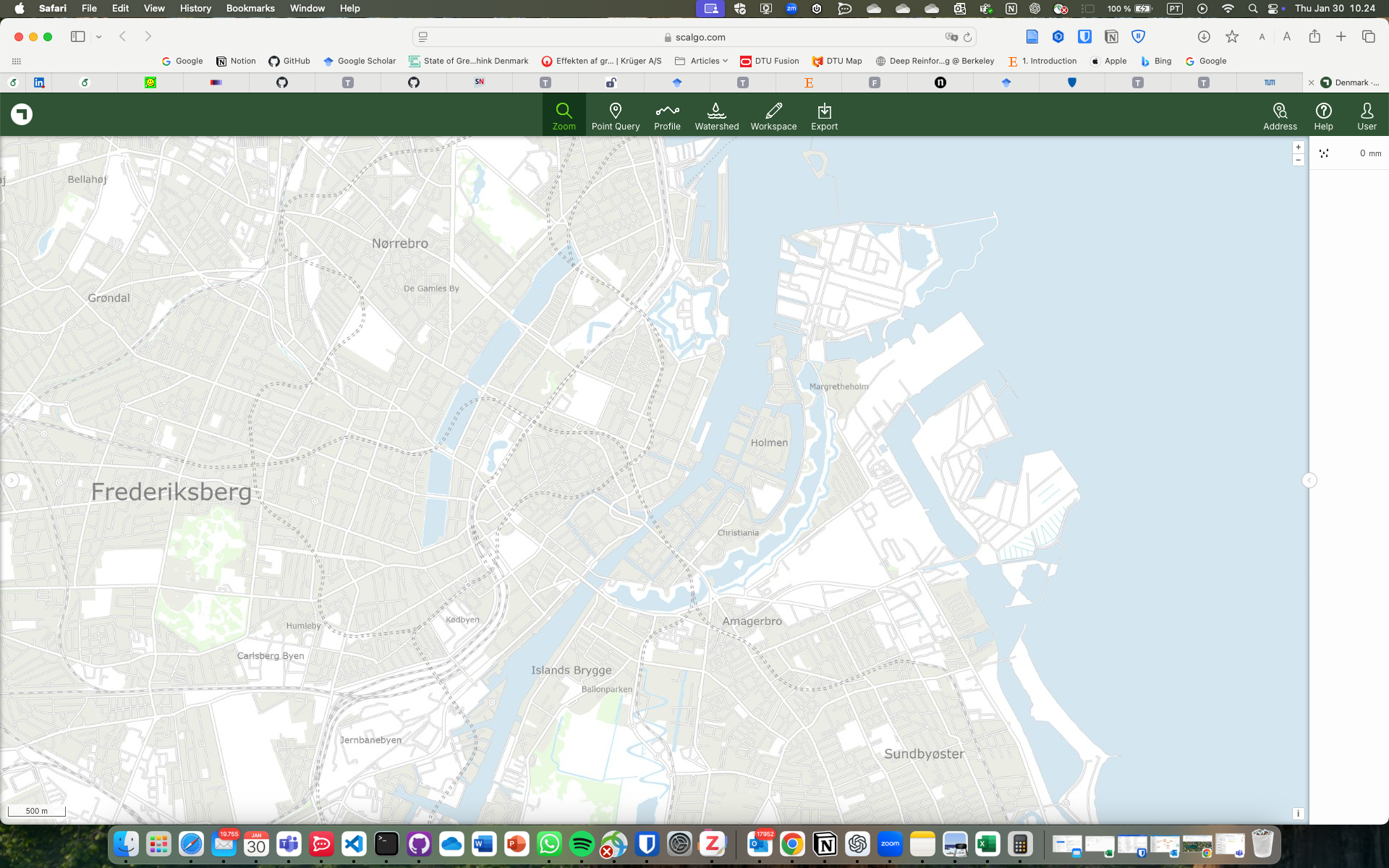} \\
        \small a) 0 mm \\
    \end{minipage} 
    \begin{minipage}[b]{.5\textwidth}
        \centering
        \includegraphics[trim={0 2.5cm 5cm 8cm},clip, width=.99\textwidth]{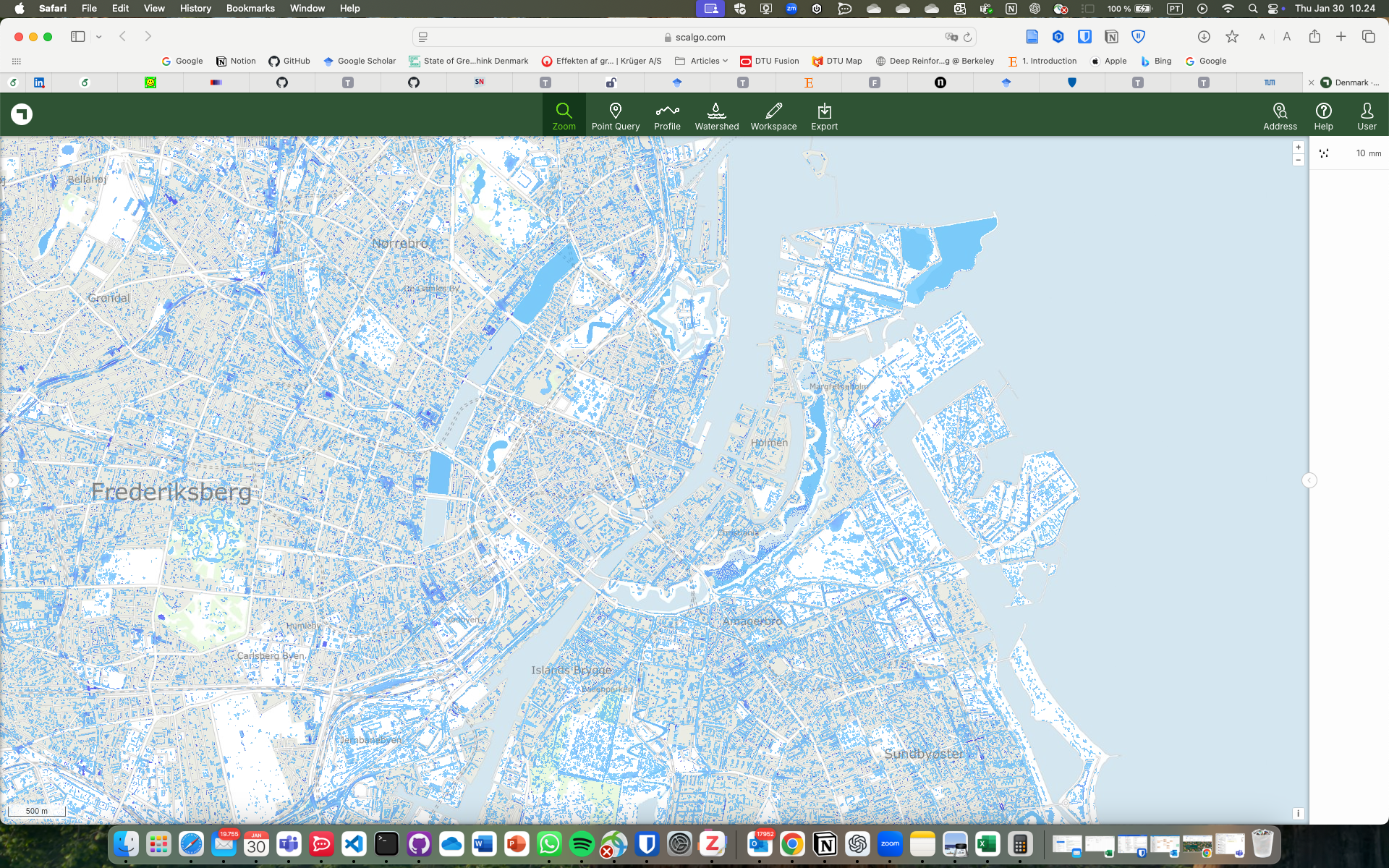} \\
        \small b) 10 mm \\
    \end{minipage} \\
    \begin{minipage}[b]{.5\textwidth}
        \centering
        \includegraphics[trim={0 2.5cm 5cm 8cm},clip, width=.99\textwidth]{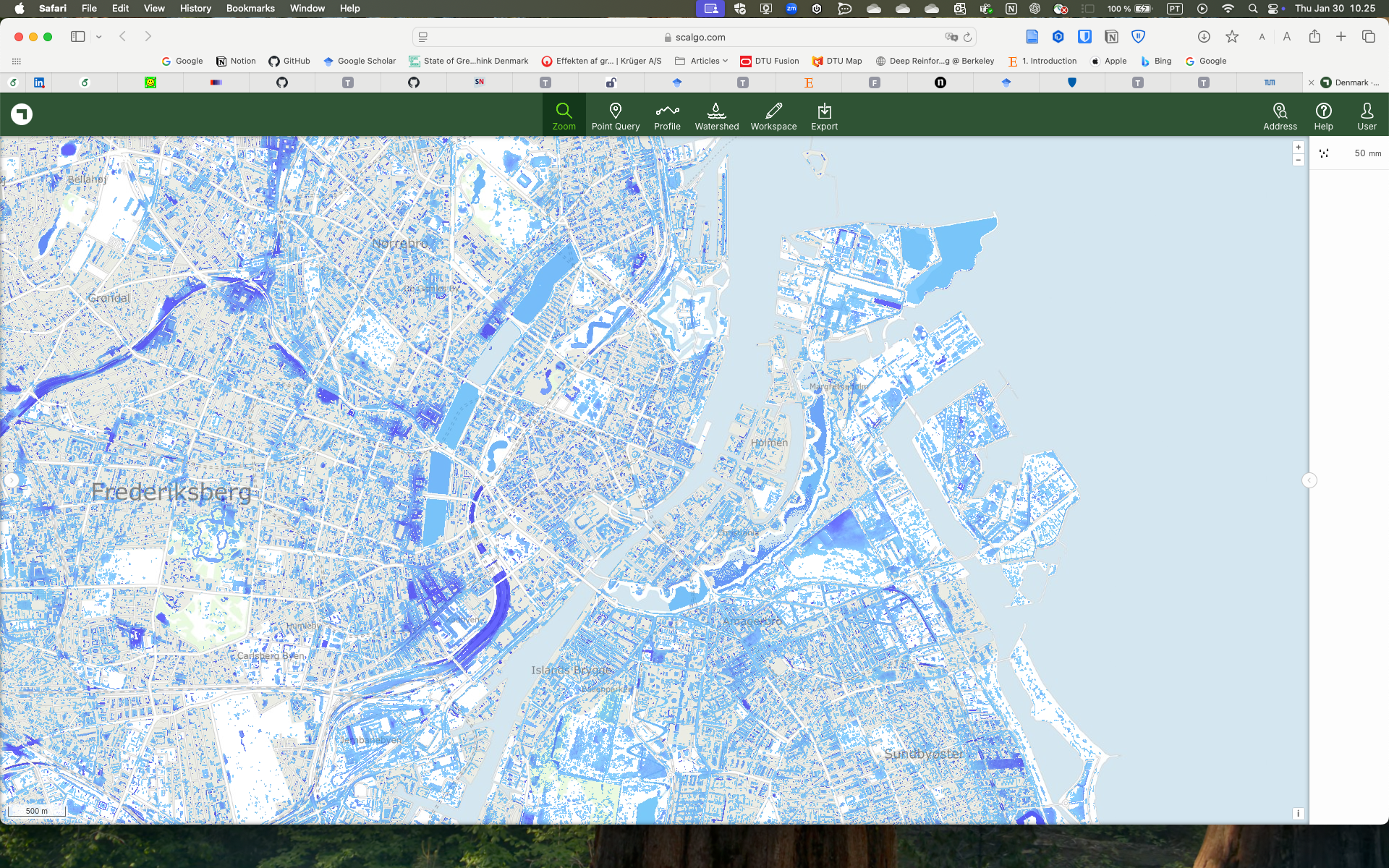} \\
        \small c) 50 mm \\
    \end{minipage}
    \begin{minipage}[b]{.5\textwidth}
        \centering
        \includegraphics[trim={0 2.5cm 5cm 8cm},clip, width=.99\textwidth]{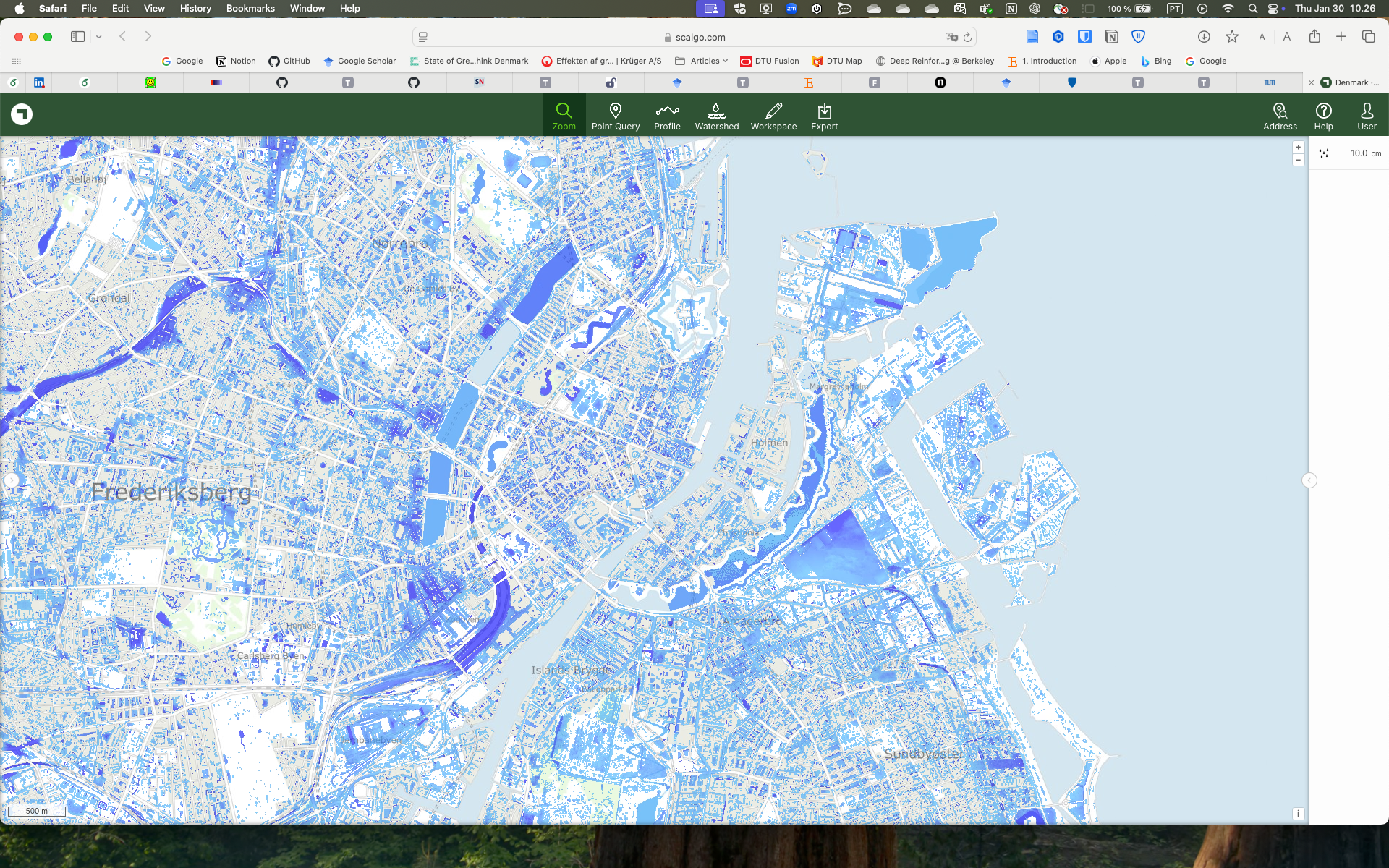} \\
        \small d) 100 mm \\
    \end{minipage} 
    \caption{Flash flood maps in Copenhagen, Denmark for rainfall intensities of 0mm, 10mm, 50mm, and 100mm. Darker blue indicates a higher water depth at the specific location. Flash flood maps were generated and retrieved from SCALGO Live \citep{scalgo}.}
    \label{fig:scalgo}
\end{figure}

This approach enables the mapping of water accumulation and the estimation of flood depth at any given location in Copenhagen for a specified rainfall event. Figure \ref{fig:scalgo} illustrates examples of simulated water accumulation for rainfall intensities of 0 mm (no rain), 10 mm, 50 mm, and 100 mm.

\subsection{Transport Accessibility}

After mapping water depth through Copenhagen for a specific rainfall event, we compute transport accessibility loss to critical infrastructure, transport-related facilities, and everyday locations using gravity-based measures versus a baseline (i.e., no flood scenario). In essence, this accessbility loss accounts for changes in travel time caused by increased water levels or untraversable roads. To map indirect impacts from water depth on travel speed, we used the functions defined by \citet{pregnolato2017impact, finnis2008field}. In essence, these modify travel speeds depending on water depths. For each transport mode specificially, travel speed is reduced incrementaly until a critical depth is reached, point at which a given segment (e.g., road) becomes untraversable. Gravity-based accessibility measures penalise locations that are further away (or take longer reach) while prioritizing those that are closer \citep{palacios2022cumulative}.

To compute accessibility, we divided Copenhagen in small hexagons (approximate radius of 50 meters). For each one, we computed the travel time to the nearest location within three predefined categories: i) critical infrastructure, including fire departments, police stations, hospitals, and other health-related facilities such as pharmacies or clinics; ii) transport-related infrastructure, including electric vehicle charging locations, gas stations, and public transit stops; and iii) everyday locations, including venues for drinking (cafes, bars, pubs), dining (restaurants, fast food outlets, food courts), recreation (sports facilities, parks), education (schools), and shopping (supermarkets). For each area’s centroid, the shortest travel time to each of these locations was computed, forming the basis for our accessibility analysis. In parallel, we compute the same accessibility measures for the baseline case, to which we compute the associated accessibility loss.

\begin{figure}[!h]
    \centering
    \includegraphics[trim={0cm 0cm 0cm 0cm},clip, width=.87\textwidth]{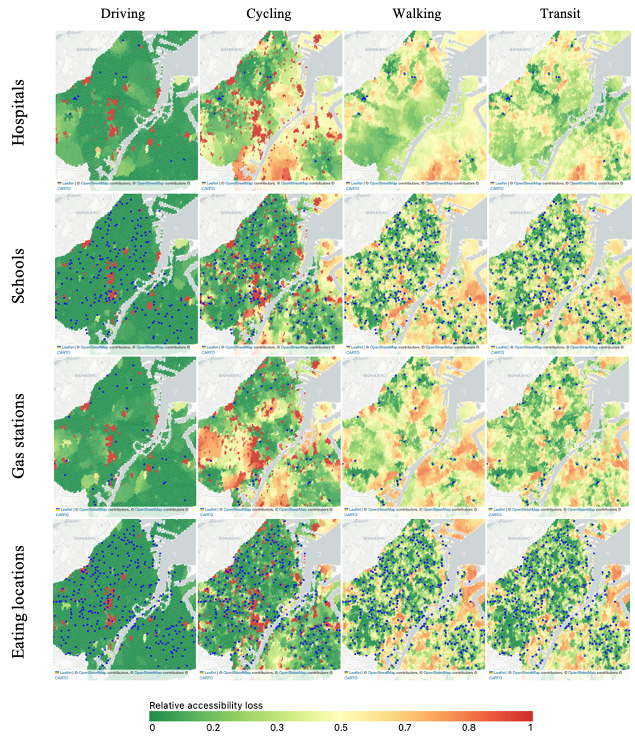}%
    \vspace{-10pt}
    \caption{Average relative accessibility loss for a 60 mm intensity rainfall event (approximately a 100-year return period) for driving, cycling, walking, and transit trips to hospitals, schools, gas stations, and eating locations.}
    \label{fig:accessibility}
\end{figure}

Accessibility loss was then taken as the average loss of three widely used gravity-based accessibility functions to measure accessibility: inverse power ($t_i^\alpha$), negative exponential ($e^{-\beta t_i}$), and modified gaussian ($e^{-t_i^2/\nu}$). These functions were applied to the estimated travel times ($t_i$) for a given trip $i$ with different impedance parameters ($\alpha, \beta, \nu$) for different transport modes to accurately account for different rates of decline. For simplicity, our analysis does not differentiate by trip purpose or consider trip-specific parameters \citet{li2018potential}. Figure \ref{fig:accessibility} showcases the relative accessibility loss for driving, cycling, walking, and transit trips in Copenhagen for a 100 mm intensity rainfall event.

\subsection{Wellbeing Modelling Framework}

For our wellbeing model we make use of the concept of motility to link changes in transport with individuals' reported life satisfaction. This approach is based on previous research linking motility with various articulations of subjective wellbeing (e.g. \cite{devosTravelSubjectiveWellBeing2013, shliselbergBehavioralFrameworkMeasuring2020}).

Motility is distinguished from more traditional metrics used in transportation research because it incorporates an individual's \textit{ability} to be mobile as well as their \textit{attitudes} towards being mobile alongside more typical metrics like accessibility. In our research, we use data gathered by the \href{https://orbit.dtu.dk/en/projects/choice-necessity-or-chance-understanding-behaviour-change-in-tran}{URGENT} project to capture these socio-cultural and psychological dimensions of motility.

The URGENT survey is a 3-year, bi-annual panel survey (\textit{n} = 12,118) that is designed to capture (among others) mobility behaviour, attitudes, norms, and constraints, alongside life and travel satisfaction in Denmark. To do this, it features a large amount of Likert-scale variables measuring respondents' agreement with a series of statements and questions relevant to the survey's remit.

\begin{figure}[!h]
    \centering
    \includegraphics[width=0.98\linewidth]{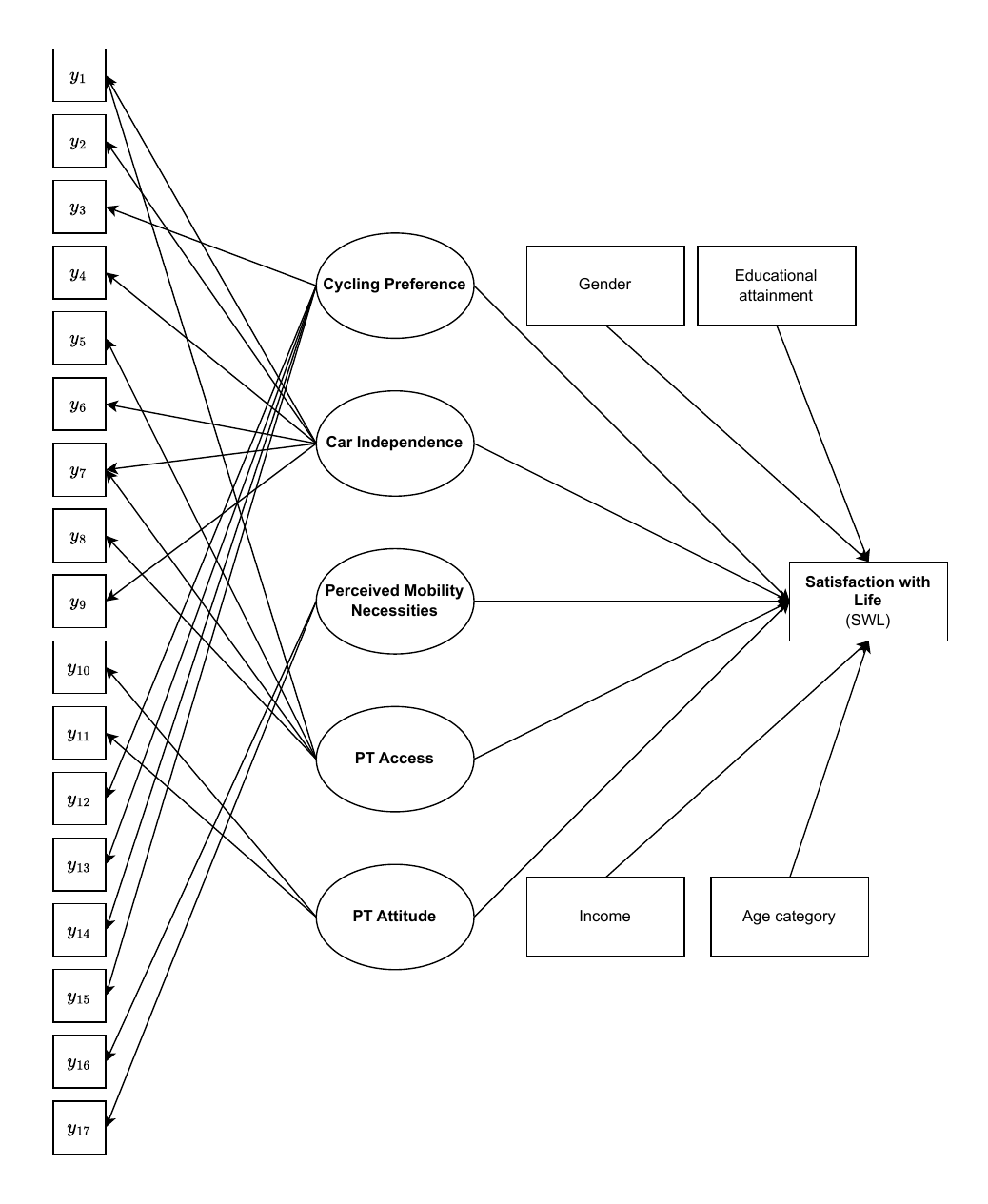}
    \vspace{-16pt}
    \caption{Preliminary wellbeing model diagram. Latent factors represented by circles, observed variables represented by rectangles. Items $y_1$ to $y_{17}$ represent the 17 survey items used for the motility components, ordered as reported in Table \ref{tab:pca_loadings}. For an in-depth look at the PCA components and their item loadings, see Table \ref{tab:pca_loadings}. For a look at the preliminary model linking the PCA components with life satisfaction, see Table \ref{tab:clm_output}.}
    \label{fig:WB_diagram}
\end{figure}

For our preliminary motility model, we include 17 survey items related to the three dimensions of motility. These are:
\vspace{-10pt}
\begin{description}
    \setlength{\itemindent}{0em} 
    \setlength{\itemsep}{0.5pt}
    \item[Access] PT access; Number of cars/household; Bike access 
    \item[Skills] Car license; PT smart card; Perceived ease of travel without car; Perceived ease of travel with PT; Perceived difficulties being mobile
    \item[Attitudes] Self-identity as car user; PT hedonic pleasure; PT privacy needs; Self-identity as cyclist; Cycle weather sensitivity; Cycle autonomy; Cycle hedonic pleasure; Perceived mobility necessities.
\end{description}
An overview of the overall model structure (from PCA to cumulative link model) can be found in Figure \ref{fig:WB_diagram}.

These survey items are then scaled and centred to turn their values into z-scores. After this, they are subjected to a Principal Component Analysis with \textit{varimax} rotation to support interpretability. Using Kaiser's Criterion ($\lambda > 1$) we retain five rotated components for analysis, which cumulatively explain 64\% of our 17 items' variance. The PCA model fits reasonably well, with the Root Mean Square Residual (SRMR) being below 0.08 (0.07). The factor loadings for each retained component, as well as component labels can be found in Table \ref{tab:pca_loadings}.

The PCA motility components are then used as covariates in a Cumulative Link Model which estimates the relationships between the five motility components, a series of socio-demographic and economic control variables, and life satisfaction. These control variables are:
\begin{itemize}
    \item \textbf{Income}: an 11-point scale ranging from less than 10.000 dkk to more than 100.000 dkk net monthly household income.
    \item \textbf{Gender}: a dummy variable indicating whether the respondent identifies as a woman or not.
    \item \textbf{Educational attainment}: Highest level of education attained, values are 'Less than Secondary', 'Vocational and Further Edcucation', and 'Tertiary Education'.
    \item \textbf{Age category}: Age category variable with a range from 18 to 80.
\end{itemize}

The dependent variable for the wellbeing model is a 5-point, self-reported Likert scale variable measuring a person's \textbf{satisfaction with life}. Preliminary results of the cumulative link model, which describes the relationship between the motility components and life satisfaction, can be found in Table \ref{tab:clm_output}. Cumulative link models are a class of ordinal regression model.

\begin{table}[!htb]
    \centering
    \small
    \scalebox{0.92}{
    \begin{tabular}{p{3.5cm}*{5}{>{\centering\arraybackslash}p{1.8cm}}}
        \toprule
        \textbf{Item} &
        \textbf{Cycling Preference (20\%)} &
        \textbf{Car Independence (16\%)} &
        \textbf{Perceived Mobility Necessities (10\%)} &
        \textbf{PT Access (10\%)} &
        \textbf{PT Attitude (8\%)} \\ 
        \midrule
        \multicolumn{6}{l}{\textbf{Access}} \\
        PT access & 0.11 & \textbf{0.25} & -0.05 & \textbf{0.67} & 0.21 \\
        Number of cars/household & -0.17 & \textbf{-0.71} & 0.11 & -0.18 & 0.00 \\
        Bike access & \textbf{0.54} & -0.22 & 0.00 & 0.27 & -0.10 \\ 
        \midrule
        \multicolumn{6}{l}{\textbf{Skills}} \\
        Car license & 0.13 & \textbf{-0.71} & -0.06 & 0.13 & 0.02 \\
        PT smart card & 0.04 & 0.13 & -0.03 & \textbf{0.52} & 0.12 \\
        Perceived ease of travel without car & 0.34 & \textbf{0.67} & -0.22 & 0.30 & 0.05 \\
        Perceived ease of travel with PT & 0.13 & \textbf{0.48} & -0.15 & \textbf{0.54} & 0.23 \\
        Perceived difficulties being mobile & -0.15 & 0.18 & -0.02 & \textbf{-0.60} & 0.19 \\ 
        \midrule
        \multicolumn{6}{l}{\textbf{Attitudes}} \\
        Self-identity as car user & -0.25 & \textbf{-0.69} & 0.13 & -0.14 & -0.08 \\
        PT hedonic pleasure & 0.06 & 0.18 & 0.05 & 0.18 & \textbf{0.75} \\
        PT privacy needs & -0.11 & 0.11 & 0.15 & -0.01 & \textbf{-0.77} \\
        Self-identity as cyclist & \textbf{0.92} & 0.20 & -0.02 & 0.08 & 0.10 \\
        Cycle weather sensitivity & \textbf{0.84} & 0.25 & -0.01 & 0.03 & 0.05 \\
        Cycle autonomy & \textbf{0.66} & 0.37 & -0.16 & 0.22 & 0.09 \\
        Cycle hedonic pleasure & \textbf{0.88} & 0.03 & -0.01 & 0.04 & 0.14 \\
        Life organisation requires mobility & -0.05 & -0.15 & \textbf{0.89} & -0.05 & -0.04 \\
        Obligations require mobility & -0.02 & -0.07 & \textbf{0.90} & -0.04 & -0.06 \\ 
        \bottomrule
    \end{tabular}}
    \caption{PCA loadings with categorised items. Loadings with absolute values greater than 0.4 in bold.}
    \label{tab:pca_loadings}
\end{table}

\begin{table}[!htb]
\centering
\small
\scalebox{0.92}{
\begin{tabular}{lrrrr}
\toprule
\textbf{Variable} & \textbf{Estimate} & \textbf{Std. Error} & \textbf{z-value} & \textbf{Significance} \\
\midrule
Cycling Preference      &  0.22 & 0.04 &  5.75 & *** \\
Car Independence         & -0.34 & 0.04 & -7.87 & *** \\
Perceived Mobility Necessities    &  0.09 & 0.04 &  2.46 & *  \\
PT Access                &  0.23 & 0.04 &  5.75 & *** \\
PT Attitude          &  0.23 & 0.04 &  5.76 & *** \\
Income                        &  0.14 & 0.02 &  8.19 & *** \\
Gender                        &  0.21 & 0.08 &  2.74 & **  \\
Educational attainment: Vocational and Further Education &  0.21 & 0.13 &  1.60 &    \\
Educational attainment: Tertiary Education &  0.24 & 0.13 &  1.83 & .   \\
Age category: 25-29              &  0.05 & 0.22 &  0.22 &     \\
Age category: 30-34              &  0.25 & 0.22 &  1.14 &     \\
Age category: 35-39              &  0.04 & 0.19 &  0.21 &     \\
Age category: 60-69              &  0.53 & 0.21 &  2.54 & *   \\
Age category: 70-80              &  0.52 & 0.23 &  2.23 & *   \\
\bottomrule
\end{tabular}}
\caption{Cumulative link model outputs, representing the relationship between motility components and self-reported life satisfaction, alongside socio-demographic and economic control variables. Significance codes: *** $p < 0.001$, ** $p < 0.01$, * $p < 0.05$, . $p < 0.1$.}
\label{tab:clm_output}
\end{table}

\clearpage\subsection{Reinforcement Learning}

Given the above IAM framework, we seek to identify the best sequence of adaptation measures that maximise individuals' wellbeing (as defined above). Reinforcement learning is a sub-field of machine learning that uses an agent-based approach learn what is the best action (adaptation measure) to take at any given time step in order to maximise a pre-defined cumulative reward function (wellbeing) \citep{sutton2018reinforcement}. By default, the environment is defined as a Markov Decision Process \citep{bellman1957markovian}, where each state is independent.

Albeit many adaptation measures can be devised, in this first work, we define four possible adaptation measures: increase street drainage, implement permeable pavements, implement an early warning system, and increase emergency services and critical infrastructure resources. At each time step, our RL agent takes one of these actions on a specific Copenhagen area (equivalent to the ones defined in the transport model) and and collects information about its state (e.g., precipitation event, period of time, direct and indirect impacts per zone, water depths on roads). In the end, this trial-and-error approach can effectively learn the best set of actions to take over time and space to maximise wellbeing.

Currently, early experiments are being run with our framework to identify good long term policies. Current efforts are focusing on comparing state of the art learning algorithms and comparing results to a set of baselines to determine the best framework and hyperparameters that yield the highest discounted sum of rewards. Currently, we are testing two RL learning algorithms: Proximal Policy Optimization (PPO) \citep{ppo2017} and Importance Weighted Actor-Learner Architectures (IMPALA) \citep{impala2018}. Each one of these has its own strengths. PPO is a widely used algorithm that serves as an initial established base algorithm. The IMPALA algorithm, however, is applied to future-proof our project. As we develop the IAM to become more realistic, increase the number of actions and their complexity, and increase our case study's area, we will use IMPALA together with distributed learning via the RLLib Python package \citep{rllib2018} to overcome computing complexity and achieve robust results in a policy-relevant timeframe.


\end{document}